%% file: main.tex
\documentclass[10pt,twocolumn,letterpaper]{article}

\usepackage{iccv}              
\usepackage{comment}
\usepackage[accsupp]{axessibility}  
\usepackage{color}
\usepackage{amsfonts}
\usepackage{amsmath}
\usepackage{amssymb}
\usepackage{colortbl}
\usepackage{eqparbox}
\usepackage{arydshln}
\usepackage{multirow}
\usepackage{stackengine}
\usepackage{soul}
\usepackage[dvipsnames]{xcolor}
\newcommand\xrowht[2][0]{\addstackgap[.5\dimexpr#2\relax]{\vphantom{#1}}}
\newcommand{\modelname}{PAR\xspace}
\newcommand{\frameworkname}{Pruning All-Rounder\xspace}

\definecolor{mygray}{gray}{0.93}
\definecolor{mygreen}{RGB}{93,173,85}
\definecolor{midblue}{HTML}{6ca6cd}

\definecolor{midred}{HTML}{d03d41}


\definecolor{iccvblue}{rgb}{0.21,0.49,0.74}
\usepackage[pagebackref,breaklinks,colorlinks,allcolors=iccvblue]{hyperref}

\def\paperID{8053} 
\def\confName{ICCV}
\def\confYear{2025}

\title{Pruning All-Rounder: Rethinking and Improving Inference Efficiency for Large Vision Language Models}

\author{Wei Suo$^{1}$, Ji Ma$^{1}$, Mengyang Sun$^{1}$, Lin Yuanbo Wu$^{2}$,
 Peng Wang$^{1}$\footnotemark[3] , Yanning Zhang$^{1}$
\\
$^1$Northwestern Polytechnical University, China. \\
$^2$Swansea University, United Kingdom. \\
{\tt\small \{suowei1994,maji,sunmenmian\}@mail.nwpu.edu.cn
\tt\small \{peng.wang,ynzhang\}@nwpu.edu.cn
}}

\begin{document}
\maketitle

\renewcommand{\thefootnote}{\fnsymbol{footnote}} 
\footnotetext[3]{Corresponding authors.} 

\input{sec/0_abstract}    
\input{sec/1_intro}   
\input{sec/2_related_work}
\input{sec/3_preliminary}

\input{sec/4_method}

\input{sec/5_experiment}
\input{sec/6_conclusion}

{
    \small
    \bibliographystyle{ieeenat_fullname}
    \bibliography{main}
}

\end{document}

%% file: sec/0_abstract.tex
\begin{abstract}

Although Large Vision-Language Models (LVLMs) have achieved impressive results, their high computational costs pose a significant barrier to wide application. To enhance inference efficiency, most existing approaches can be categorized as parameter-dependent or token-dependent strategies to reduce computational demands. However, parameter-dependent methods require retraining LVLMs to recover performance while token-dependent strategies struggle to consistently select the most relevant tokens. In this paper, we systematically analyze the above challenges and provide a series of valuable insights for inference acceleration. Based on these findings, we propose a novel framework, the Pruning All-Rounder (PAR). Different from previous works, PAR develops a meta-router to adaptively organize pruning flows across both tokens and layers. With a self-supervised learning manner, our method achieves a superior balance between performance and efficiency.  Notably, PAR is highly flexible, offering multiple pruning versions to address a range of acceleration scenarios. The code for this work is publicly available at \href{https://github.com/ASGO-MM/Pruning-All-Rounder}{https://github.com/ASGO-MM/Pruning-All-Rounder}.

\end{abstract}

%% file: sec/1_intro.tex
\section{Introduction}
\label{sec:intro}

Substantial progress has been made in the development of Large Vision-Language Models (LVLMs) in recent years \cite{llava, llava1-5, qwen-vl, instructblip,suo2024rethinking,ma2024c3,suo2025octopus,suo2023s3c}.
Despite being able to handle various Vision-Language (VL) tasks, the applications of these models become restricted due to the increasing computational costs. In fact, the significant computational burden of LVLMs originates from billions of model parameters and lengthy input sequences. Thus, there are two paradigms to address such a problem: parameter-dependent and token-dependent compression strategies.

\begin{figure}[t]
\centering
\includegraphics[width=0.5\textwidth]{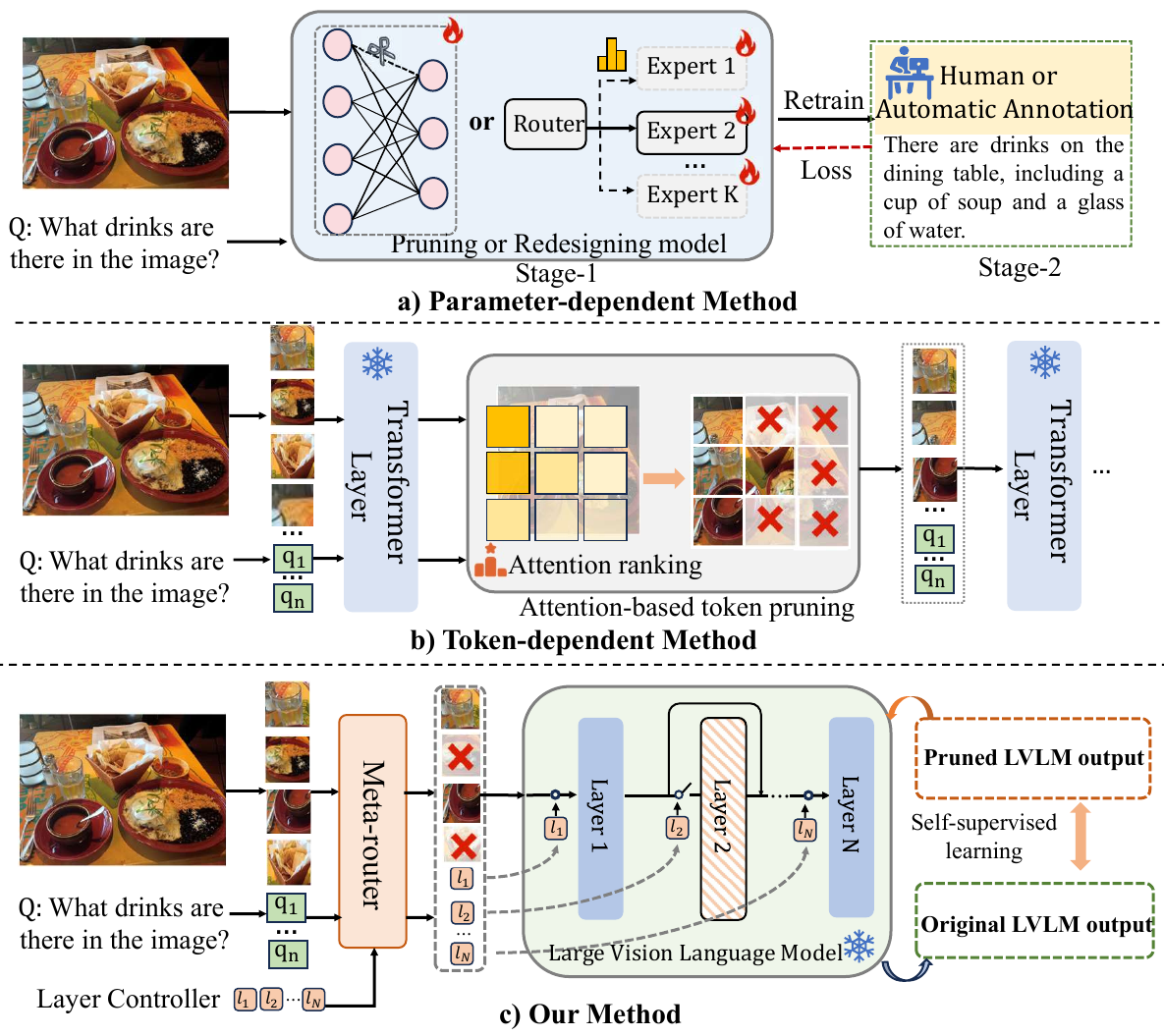}
\caption{Comparison of different model compression methods. (a) The parameter-dependent strategy requires an additional fine-tuning process to bridge the performance gap. (b) The token-dependent pruning method accelerates inference by reducing the input length. (c) Our method focuses on jointly compressing the inputs and parameters in a self-supervised manner.}
\label{fig:fig1}
\end{figure}

As shown in Fig.\ref{fig:fig1}(a), the parameter-dependent acceleration approaches mostly follow an inherent two-stage paradigm to enhance the inference efficiency. In practice, they first require localizing irrelevant weights~\cite{Upop,multiflow,lottery_ticket_vlm} or redundant structures~\cite{dynamicrouting,RoE_LLaVA,stream_layer} within LVLMs. 
Then, these pruned or newly designed models are re-trained to minimize performance degradation. On the other hand, as shown in Fig.~\ref{fig:fig1}(b), the token-dependent pruning technologies compress inference costs by reducing the model's input length~\cite{tome, dynamic_vit, fastv, hired, llavaprumerge}. Considering that not all visual tokens are related to a given question, this approach usually applies attention scores as a criterion to discard~\cite{fastv} or merge non-significant tokens~\cite{llavolta}. 

Although essential progress has been made,
they face the following limitations:
1) Parameter-dependent acceleration methods mostly involve an additional procedure to fine-tune the pruned model~\cite{RoE_LLaVA,stream_layer} or retrain a new VL model from scratch~\cite{multiflow,lottery_ticket_vlm}. This complex pipeline requires substantial annotations and huge computational resources to update the parameters of LVLMs.
2) For a long visual input, even if each token only has two options (\textit{i.e.,} keep or discard), the solution space is still vast due to combination explosion~\cite{patch_rank}. How to select the most relevant tokens is still a significant challenge.
3) Despite parameters and inputs jointly leading to oversized computational costs, previous efforts treat them as two independent research lines and ignore their relationship.

In this paper, we systematically analyze the above limitations (a detailed discussion can be found in Sec.~\ref{sec:preliminary}) and introduce new insights as follows: 1) We extend the layer redundancy study~\cite{shortgpt} to the multimodal domain and find that the scale of LVLMs can be effectively reduced by directly skipping some redundant layers. This insight suggests a promising parameter compression strategy  without complex procedures to update LVLMs' weights after pruning. 
2) With regard to token pruning, extensive results show that these rule-based token selection methods (\emph{e.g.,} attention score ranking~\cite{fastv}) 
cannot consistently localize the most relevant tokens. 
Inspired by~\cite{vcd}, we add noise to rule-based token selection results and empirically show that some better token combinations can be captured by expanding selection space with noise.
3) We experimentally demonstrate that tokens and layers are correlated. Therefore, 
building a joint optimization model is crucial for integrating the pruning process.

Based on these insights, we argue that there is still a substantial opportunity to improve model efficiency, particularly for models already deployed. Consequently, we introduce a new framework called \textbf{P}runing \textbf{A}ll-\textbf{R}ounder, abbreviated as \modelname. As illustrated in Fig.\ref{fig:fig1} (c), rather than relying on a single strategy to accelerate inference, our \modelname focuses on jointly compressing redundant layers and tokens with self-supervised learning. Specifically, we first develop a simple yet effective transformer-based block as a meta-router to adaptively organize the pruning workflow. Additionally, considering that the traditional rule-based token selection method~\cite{fastv} can only offer a limited reference, we employ a noise perturbation mechanism to expand the search space. Finally,
without depending on human or automated labeling, the meta-router can be optimized by contrasting changes in model responses.
Benefiting from the above designs, \modelname not only achieves a better trade-off between performance and computational costs but also offers several potential advantages: \textbf{ 1) Convenience.}
Our method effectively compresses the computational scales without updating LVLMs' parameters after pruning or relying on extensive human-labeled annotations, making it conveniently applicable to already deployed models. \textbf{2) Model Agnostic.} Without bells and whistles, the \modelname can be regarded as a plug-and-play module and easily integrated into other frameworks.
\textbf{3) Flexibility.} Different domains have distinct pruning preferences. For example, discarding tokens would be risky in medical scenarios~\cite{trustworthy_healthcare}. As an all-rounder, our method provides users with multiple selectable versions.
Our contributions are as follows:

1) We systematically discuss the redundancy of LVLMs at both the layer and token levels. By conducting qualitative and quantitative analyses, our work draws a series of meaningful insights for inference acceleration.

2) We propose a new framework \frameworkname to reduce the computational costs of LVLMs in a self-supervised manner. Different from the previous works, the \modelname can jointly compress the parameters and tokens without re-adjusting the model's weights after pruning. 

3) The proposed \modelname introduces multiple alternative layouts to reduce the FLOPs while maintaining the original capacity of models. Extensive experiments show the effectiveness of our method across nine different benchmarks.

%% file: sec/2_related_work.tex
\section{Related work}
\label{sec:related_work}

\subsection{Large Vision-Language Models}

In recent years, Large Vision-Language Models (LVLMs) have attracted significant attention due to their powerful reasoning capabilities. 
Unlike previous specialized vision-language models, these LVLMs inherit the capabilities of language models and can follow different human instructions. Typically, these models~\cite{llava,llava1-5,qwen-vl,instructblip} consist of three main modules: Visual Encoder, Large Language Model (LLM), and Multimodal Connector. However, with the continuous expansion of model scale, they are constrained by high computational costs. 
In this paper, we aim to accelerate the reasoning process of LVLMs to facilitate more user-friendly models.
Meanwhile, since LLMs constitute the majority of parameters in LVLMs~\cite{minivlm,distillvlm,Upop}, our work focuses on compressing the LLM component to enhance overall efficiency.

\begin{figure*}[t]
  \centering
  \begin{subfigure}[t]{0.325\linewidth}
\includegraphics[width=\textwidth,height=4cm]{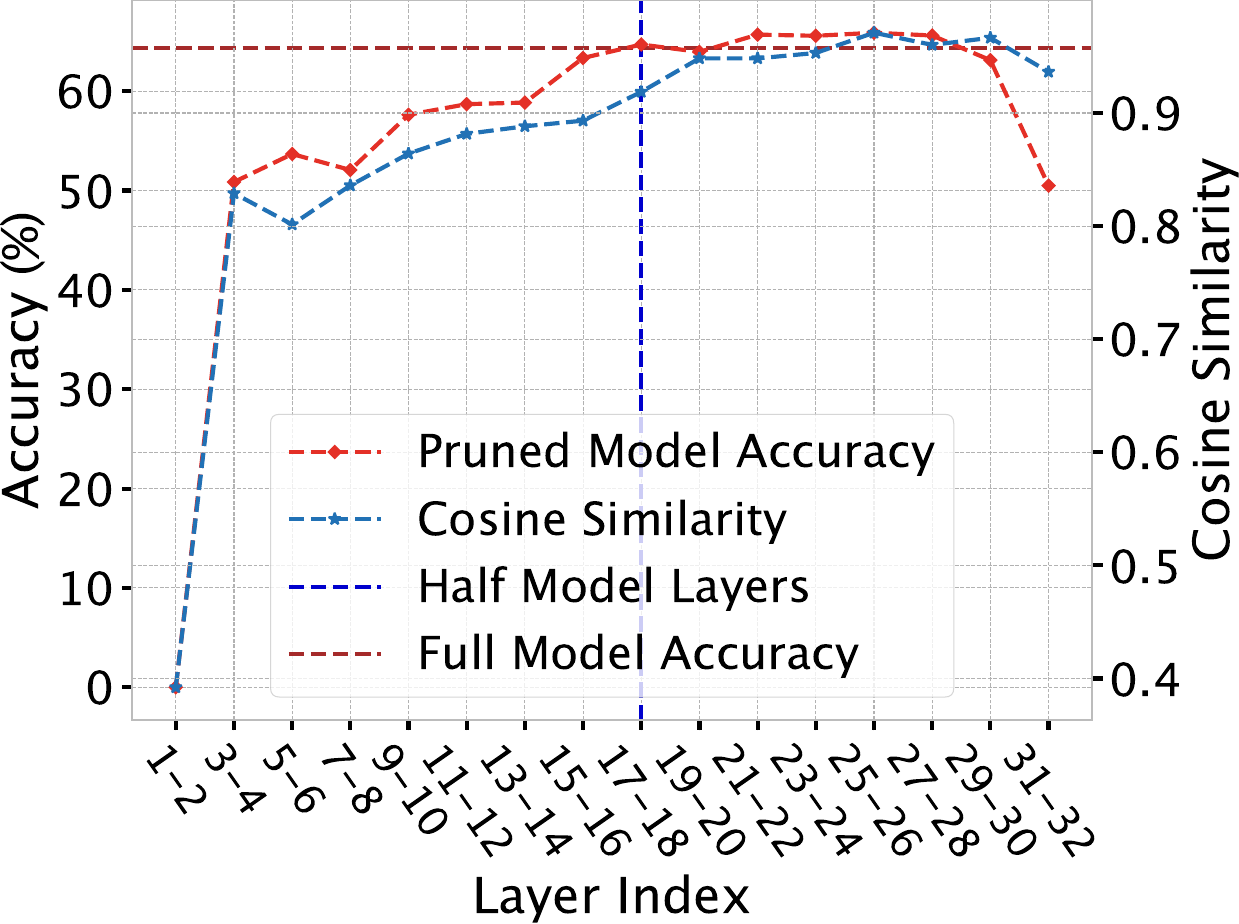}
    \caption{}
    \label{fig2:sub1}
  \end{subfigure}
  \hfill
  \begin{subfigure}[t]{0.31\linewidth}
    \includegraphics[width=\textwidth,height=4cm]{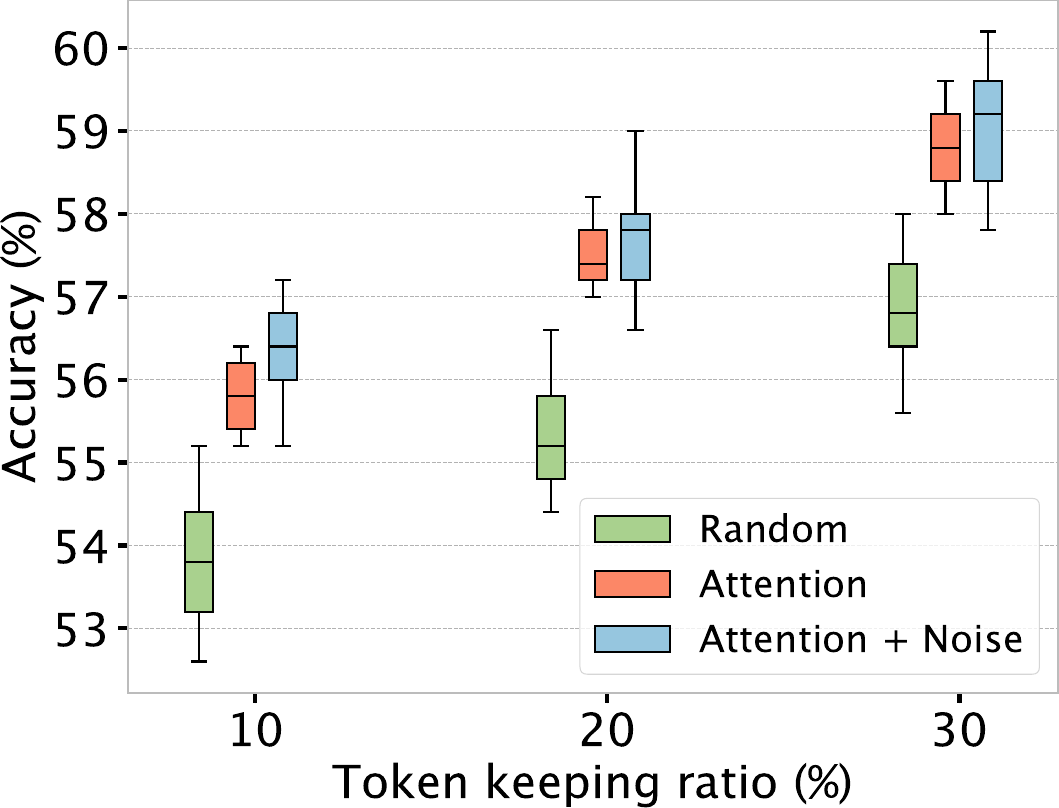}
    \caption{}
    \label{fig2:sub2}
  \end{subfigure}
  \hfill
  \begin{subfigure}[t]{0.31\linewidth}
\includegraphics[width=\textwidth,height=4.02cm]{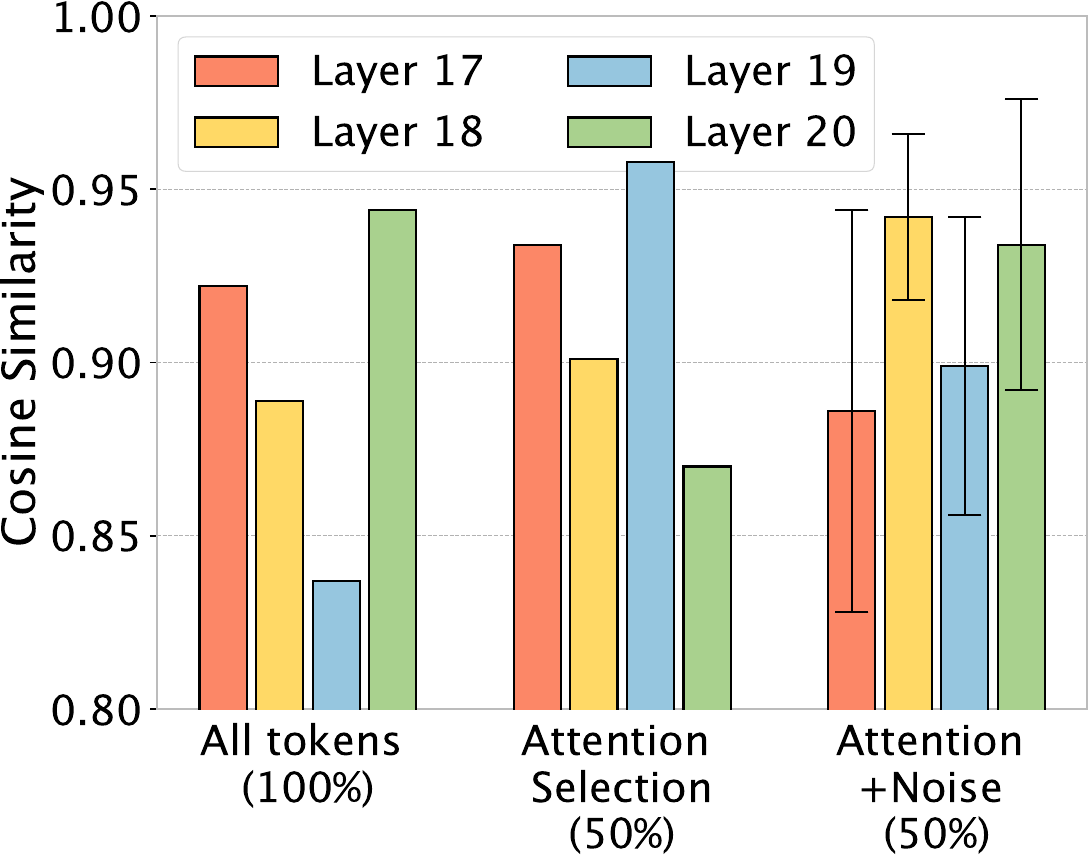}
    \caption{}
    \label{fig2:sub3}
  \end{subfigure}
  \caption{Preliminary experiments of LVLMs: (a) The layer-level redundancy analysis. We observed that some redundant layers can be directly skipped without significantly affecting inference performance. (b) The token-level redundancy analysis. Some better token combinations can be captured by introducing noise. (c) The changes in cosine similarity across layers and tokens. The results show that layers and tokens are correlated.}
  \label{fig:preliminary}
\end{figure*}

\subsection{Parameter Pruning}

Many researchers have attempted to reduce the computational costs of LVLMs by decreasing model scales, which has led to the development of parameter pruning methods~\cite{minivlm, distillvlm,Upop, vlm_lottery_tickets, multiflow}. Existing parameter pruning approaches for VLMs can be broadly categorized into two types: task-specific pruning~\cite{minivlm, distillvlm} and task-agnostic pruning~\cite{Upop, vlm_lottery_tickets, multiflow}. Given that prevalent LVLMs can handle a diverse array of vision-language tasks, task-agnostic pruning appears more practical than task-specific approaches. However, these methods~\cite{Upop,multiflow,vlm_lottery_tickets} typically involve complex weight search algorithms and require an additional retraining stage after pruning, which limits their practicality, particularly for already deployed LVLMs.
Meanwhile, several recent efforts have focused on redesigning model architectures, such as Mixture-of-Experts (MoE)~\cite{RoE_LLaVA} and Mixture-of-Depth (MoD)~\cite{luo2024gamma}, to reduce computational burden. 
Unlike these approaches, we demonstrate the significant layer redundancy present in current LVLMs, and remove redundant layers without model retraining while preserving the original capabilities.

\subsection{Token Pruning}

One well-known drawback of the Transformer~\cite{transformer} architecture is its quadratic computational complexity with respect to input length. Therefore, lengthy visual inputs represent another key factor impacting model efficiency. To accelerate LVLMs from the input perspective, researchers have proposed pruning~\cite{vtw,fastv,hired,g-prune,fit-prune} or merging~\cite{tome,adaptive_token_Vit,llavaprumerge} unnecessary tokens.
For example, FastV~\cite{fastv} is the first work to identify redundant tokens through attention score ranking. VTW~\cite{vtw} analyzes attention scores between different token types (\textit{i.e.,} instruction, visual, and text tokens) and discovers that visual tokens in deeper layers are rarely attended to, thus simply pruning all visual tokens in these layers. Using a slightly different attention-based approach, LLaVA-PruMerge~\cite{llavaprumerge} and HiRED~\cite{hired} visualize attention scores between CLS tokens and spatial visual tokens. Their results show that many attention values are near-zero, revealing the redundancy of numerous visual tokens. More recently, Fit-Prune~\cite{fit-prune} prunes tokens by minimizing attention distribution divergence between pre-pruning and post-pruning states.
Different from the above methods, our work aims to explore an easier-to-deploy approach that jointly compresses redundant layers and tokens without updating the LVLMs' parameters after pruning.

%% file: sec/3_preliminary.tex
\section{Preliminary}
\label{sec:preliminary}

\subsection{Large Vision-Language Models }
\label{sec:Sec3.1}
Given an image $I$, LVLMs can perform multiple Vision-Language tasks by following the user's textual instruction $Q$. To accomplish the corresponding reasoning task, image input $I$ would first be encoded into token sequence [$v_1,v_2,...,v_s$] by a visual encoder (\emph{e.g.,} CLIP-ViT~\cite{clip}) and a multi-modal connector (\emph{e.g.,} MLP~\cite{llava1-5} or Q-former~\cite{instructblip}), where $s$ denotes the sequence length of image tokens. Meanwhile, the textual input $Q$ is converted as corresponding language embedding [$q_1,q_2,...,q_c$], where $c$ is the length of text input. Finally, the response of $t$-th generative step can be obtained in an auto-regressive manner:

\begin{equation}
    p(y_t) = \prod \limits_{t=1}^Lp_\theta(y_t|{v_1},...{v_s},q_1,...,q_c,y_{<t}).
    \label{eq:1}
\end{equation}
Here, we omit the systematic prompt tokens for simplicity. 
In general, the above generation process can also be formalized as $y = f(x,\theta)$,
where $y$, $x$, and $\theta$ are outputs, inputs and parameters of LVLMs, respectively. The $f$ denotes the mapping from inputs to outputs.
It can be seen that the high computational costs of LVLMs mainly arise from calculations with $\theta$ and $x$. Therefore, two research paradigms are presented for model acceleration: parameter-dependent and token-dependent strategies.

However, several issues remain to be answered before accelerating the LVLMs: 1) Existing parameter-dependent acceleration methods need to fine-tune pruned LVLMs to bridge the performance gap~\cite{vlm_lottery_tickets, multiflow}, which limits their feasibility of compressing already deployed models. 2) Previous works have proved that not all visual tokens are related to inference~\cite{fastv}, but how to select the most relevant tokens is still an open question. 3) The $\theta$ and $x$ jointly lead to vast computational costs, while their relationship has yet to be thoroughly evaluated. 

Next, we aim to answer the above questions with the popular VL model LLaVA-1.5-7B~\cite{llava}. Meanwhile, to ensure that the conclusions are generalizable, related experiments are established by combining five benchmarks: ScienceQA~\cite{scienceqa}, AOKVQA~\cite{aokvqa}, POPE~\cite{pope}, SEED-Bench-Image~\cite{seed_bench} and MM-Bench~\cite{mmbench}. 

\subsection{Layer Redundancy }
\label{sec:Sec3.2}
As mentioned above, these parameter-dependent compression strategies involve an additional process that updates the parameters of pruned LVLMs to minimize performance degeneration~\cite{lottery_ticket_vlm,layer_skipping_1,layer_skipping_3}, which is unfriendly for practical applications. In this section, we aim to explore a more \textit{easy-to-deploy} approach for parameter compression. Since existing LVLMs are composed of a stack of transformer blocks, we focus on these basic computational units (\emph{i.e.,} layers). 

Though the field of NLP has proved LLMs can directly short-cut some redundant layers~\cite{shortgpt,unreasonable_ineffectiveness}, the related study is still limited in our community. Inspired by~\cite{layer_skipping_1,layer_skipping_3}, we conduct related experiments to quantitatively analyze the layer-wise redundancy of LVLMs. In particular, as shown in Fig.~\ref{fig:preliminary}(a), 
given the above five benchmarks, we first compute the averaged cosine similarity of the hidden states between two adjacent layers. 
The results show that these adjacent layers exhibit high similarity, meanwhile, the last $16$ layers have higher redundancy (\emph{i.e.,} the blue line).
Correspondingly, the reasoning performance (\emph{i.e.,} the red line) shows a limited degradation when skipping these latter layers.
More importantly, \textbf{\textit{the layer-level redundancy phenomenon offers a promising parameter compression strategy due to avoiding additional training process.}}

\subsection{Token Redundancy }
\label{sec:Sec3.3}
In this section, we focus on lengthy visual input which is the other important reason for high computational costs. Considering that not all visual tokens are relevant to a given textual input, existing token selection approaches mostly apply attention score ranking to select the token subset~\cite{fastv, hired, llavaprumerge}. 
For example, FastV~\cite{fastv} uses attention scores from the second layer to select tokens. However, we argue that such a rule-based method is incapable of consistently selecting the most relevant tokens in the vast solution space.

To validate this hypothesis, we utilize the attention ranking in FastV~\cite{fastv} as a reference and add the Gaussian noise $\mathcal{N}\sim(0,10)$ with 5 different seeds to explore whether better token combinations exist.
As shown in Fig.~\ref{fig:preliminary} (b), we compute accuracies with three token pruning strategies (\emph{i.e.,} random selection, attention ranking, and perturbed attention ranking) under different token preserving ratios (\emph{i.e.,} 10\%, 20\% and 30\%). From the red box (attention ranking) and the green box (random selection), we find that attention-based token selection methods outperform random dropping by a large margin. 
However, by adding Gaussian noise~\cite{gaussian} to the attention values (the blue box), we see that the performance can be enhanced further. The above results indicate that the rule-based method can only provide a limited reference. 
Meanwhile, as a plausible approach, the noise perturbation has the potential to capture better token combinations by expanding the search space. Based on the above experiments, we conclude that \textbf{\textit{although using attention ranking provides an important pruning reference, some better token combinations can be obtained by introducing the noise perturbation.}}

\subsection{Joint Compression Discussion}
\label{sec:Sec3.4}
To further enhance inference efficiency, a natural idea is to integrate the above two pruning paradigms. In this section, we empirically investigate this strategy by computing the layer-wise cosine similarity across different token combinations. 
As we already demonstrated that deep layers exhibit higher redundancy compared to the shallow layers in Sec.~\ref{sec:Sec3.2}, the related experiments are conducted using these deep layers (\textit{i.e.,} layers after 16th).
Specifically, as shown in Fig.~\ref{fig:preliminary}(c), we first calculate the cosine scores 
from the 17th to the 20th layer
where all tokens are preserved (\emph{i.e.,} ``All tokens''). It can be observed that the 17th and 20th layers can be prioritized for removal if layer-wise redundancy is solely considered. However, when 50\% tokens are dropped using attention ranking, we observe a dramatic change in the cosine similarity of these layers. Furthermore, we add 5 randomly sampled Gaussian noise $\mathcal{N}\sim(0,10)$ to the attention scores and remove tokens based on perturbed scores. We find the cosine similarity of these layers is not fixed and the token combinations will greatly influence the ranking of layer-wise importance.
We conclude that\textbf{\textit{ the layer-dependent and token-dependent strategies are correlated, and thus a joint optimization model is crucial to integrate them.}} 

\subsection{Summary}
Below, we summarize our investigations against layer redundancy and token redundancy.

(1) For redundant parameters, we observe that directly skipping some deep layers does not dramatically affect the reasoning ability of LVLMs. This finding extend the layer redundancy study in NLP~\cite{shortgpt} to the multimodal domain and provides an \textit{easy-to-deploy} parameter compression paradigm  without additional processes to update the parameters of pruned LVLMs.

(2) Regarding the lengthy visual input, we reveal that the existing attention-based token selection methods cannot consistently select the most important tokens and lead to sub-optimal results.
Fortunately, some better token combinations can be captured by introducing the Gaussian noise.

(3) We find that the token dropping and layer skipping are correlated. Therefore, an ideal acceleration method should jointly model them to achieve a better trade-off between performance and computational costs.

%% file: sec/4_method.tex
\begin{figure*}[t]
    \centering
\includegraphics[width=1\textwidth]{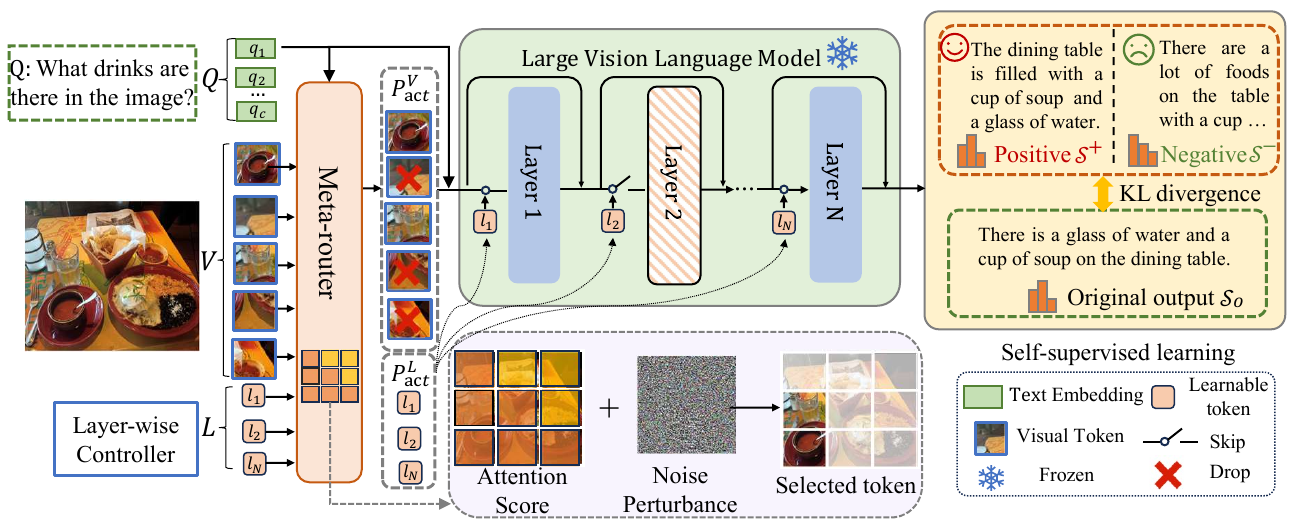}
    \caption{Overview of our Pruning All-Rounder (PAR). Given an image and corresponding question, our method focuses on applying a meta-router to adaptively prune redundant tokens and layers. Without human or automatic labeling, our PAR can be optimized by calculating the KL-divergence scores. Finally, as an all-rounder, the meta-router also provides multiple selectable pruning layouts to tackle different scenarios.}
\label{fig:model_arch}
\end{figure*}

\section{Method}
\label{sec:method}
Based on these insights, we propose a new framework Pruning All-Rounder (PAR), which can adaptively drop redundant tokens and short-cut unnecessary layers in a self-supervised manner. As shown in Fig.~\ref{fig:model_arch}, the PAR focuses on leveraging a lightweight transformer block as the meta-router to compress the high computational costs of LVLMs. More importantly, our approach can be individually optimized or implemented to tackle different scenarios. In other words, existing works such as rule-based token pruning~\cite{fastv} or early exiting mechanism~\cite{early_exiting_2} can also be regarded as specific versions of our method. Next, we will introduce our approach in detail.

\subsection{Pruning All-Rounder}

For a well-trained LVLM, 
the core objective of our work is to build a pruning flow by a meta-router $\mathcal{O_\phi}$ with parameters $\phi$.  Meanwhile, considering the complex relationship between tokens and layers, this $\mathcal{O_\phi}$ can adaptively model them and provide more flexible pruning layouts.

Specifically, given a lengthy visual input sequence $V=\{v_i\}_{i=1}^{S}$ and textual instruction $Q=\{q_i\}_{i=1}^{C}$, our work would only prune visual token $V$ due to $S\gg{C}$~\cite{fastv,vtw}. Moreover, we build a group of learnable embedding $L=\{l_i\}_{i=1}^N$ as a layer-wise controller to manage the layer-skipping operation, where $N$ is the number of candidate layers. Because Sec.~\ref{sec:Sec3.2} has proved that the last 16 layers exhibit higher redundancy, we set the $N=16$ to represent the index of these layers.
Then, the meta-router $\mathcal{O_\phi}$ is utilized to model the pruning process by feeding the $V$, $Q$ and $L$.
To ensure lightweight, our $\mathcal{O_\phi}$ is a basic transformer block~\cite{transformer} with $\sim$40M trainable parameters. The above computations can be formulated as:
\begin{equation}
    Z_{out} = \mathcal{O}_\phi(\mathrm{concat}[V;Q;L]),
\end{equation}
where $\rm{concat}$ represent concatenate operation. 
Since our work focuses on dropping redundant visual tokens and layers, the $Z_{out}=[Z_{out}^{V}; Z_{out}^{L}]$ would only collect the corresponding outputs from $V$ and $L$, while the outputs from $Q$ are omitted. 
Taking advantage of the self-attention mechanism~\cite{transformer}, each element of $Z_{out}$ can adaptively aggregate the information from other inputs. Finally, we split the $Z_{out}$ to $Z_{out}^{V}$ and $Z_{out}^{L}$, which would be mapped into the corresponding action space (\emph{i.e.,} keep or discard). We take $Z_{out}^{V}$ as an example, the ranked scores $P_{act}^V$ are obtained by:
\begin{equation}
\label{eq:3}
    Z_{act}^{V}= \sigma(\mathrm{MLP}(Z_{out}^{V})),
    P_{act}^{V} = \mathrm{Rank}(Z_{act}^{V}),
\end{equation}
where $\rm Rank$ and $\sigma$ refer to the ranking operation and sigmoid active function. MLP denotes a Multi-Layer Perceptron. For $Z_{out}^{L}$, we also use the Eq.~\ref{eq:3} to obtain $P_{act}^{L}$.
During inference, we directly select top-$K$ layers and top-$M$ tokens to discard.
Meanwhile, the pruning flow can also be customized by only ranking either $Z_{act}^{V}$ or $Z_{act}^{L}$ to tackle different scenarios. For example, users can only rank $Z_{act}^{V}$ to achieve token pruning.

\subsection{Model Optimization}
It can be noted that Eq.\ref{eq:3} involves a non-differentiable operation, meanwhile, the meta-router $\mathcal{O_\phi}$ is difficult to optimize without explicit labels. To address this challenge, we reformulate the training procedure of $\mathcal{O_\phi}$ as a preference optimization task and use the DPO technique~\cite{dpo} to train our meta-router. Furthermore, the above training process can be implemented in a self-supervised manner, allowing it to work without any human or automated labeling.
\paragraph{Preference Data Construction.} 
The core of DPO technology is to guide the model to generate expected responses by constructing high-quality positive and negative samples. In this paper, we extend this idea into the model compression domain with self-supervised learning.

In particular, given an image-question pair $(V,Q)$ in an unlabeled multi-modal dataset, we remove $K$ layers and $M$ tokens to construct contrastive pruning actions. Here, the $K$ is randomly selected from the last 16 layers based on the conclusions in Sec~\ref{sec:Sec3.2}. 
As for visual tokens, to 
explore better combinations, we apply Gaussian noise to perturb the rule-based scores ranking~\cite{fastv} as mentioned in Sec~\ref{sec:Sec3.3} and discard $M$ tokens with the lower scores. By performing the above operation 5 times for each sample, we can build a pruning action set $\mathcal{S}$. Moreover, as a meta-router, this preference set $\mathcal{S}$ can also be specifically used for layer or token pruning when the $K$ or $M$ is set to 0.
Then, we divide the $\mathcal{S}$ into positive action $\mathcal{S}^+$ and negative action $\mathcal{S}^-$ by comparing them with the original outputs $\mathcal{S}_o$ (\textit{i.e.,} outputs of the full LVLM). Considering that the outputs of the pruned model will exhibit significant changes compared to the original ones, we use KL-divergence~\cite{KL-divergence} to measure whether the reasoning is correctly executed. 
Finally, following~\cite{li2023silkie}, we use balanced positive and negative sets to construct the preference dataset. In our experiments, we observed that satisfactory results have been achieved using only 5,000 unlabelled samples.

\paragraph{Preference Optimization Process.}

Compared to previous reinforcement learning~\cite{rlhf,llava-rlhf}, the DPO uses a policy model and a reference model to avoid the complex optimization process.
Benefiting from its simplicity and stability, we apply this technology to optimize our $\mathcal{O}_\phi$.
Meanwhile, previous works~\cite{reference_free_dpo,reference_free_dpo_1} have demonstrated that better or comparable performance than the original DPO can be achieved by removing the reference model. Therefore, given the preference dataset $\mathcal{D} = \{ \mathcal{S^+}, \mathcal{S^-} \}$, our optimization objective is formulated as follows:
\begin{equation}\label{eq:reward_model}
\begin{aligned}
   \max_{\mathcal{O}_\phi} \mathbb{E}_{(x, \mathcal{S^+}, \mathcal{S^-})\sim \mathcal{D}}&\bigl[\log \sigma(\beta\log \mathcal{O}_\phi(\mathcal{S}^+|x)\\ 
   &-\beta\log\mathcal{O}_\phi(\mathcal{S}^-|x) ],
\end{aligned}
\end{equation}
where $x=(V,Q,L)$ represents the input sequence. We follow~\cite{reference_free_dpo,reference_free_dpo_1} and set the $\beta$ as 1. With the above optimization objective, the \modelname can learn to select the suitable model acceleration strategy based on different inputs. Note that we only optimize the lightweight $\mathcal{O}_\phi$, while the parameters of LVLM would be frozen. More importantly, the $\mathcal{O}_\phi$ is only optimized to determine which visual tokens (or model layers) to prune \textbf{without changing the features of visual tokens or performing re-alignment between vision and language.}
By keeping the parameters of LVLMs frozen, we can enhance the efficiency of LVLMs with minimal training overhead (\textit{e.g.,} approximately 10 minutes of training time on a single NVIDIA 3090 GPU).

%% file: sec/5_experiment.tex
\section{Experiment}
\label{sec:experiment}
    
\begin{table*}[t]
\centering
\small
\caption{
For simplicity, the \textit{T} and \textit{L} represent the number of tokens and layers that are \textbf{preserved}. $\dag$ denotes our meta-router is trained on this setting and other results are directly tested.
$^*$ indicates that the original paper did not provide the results, and they are reproduced under the same settings. The results are sorted in descending order of TFLOPs. Note that the best results within the same TFLOPs range are highlighted in \textbf{bold}. }
    \begin{tabular}{lccccccccc}
         \hline
         \xrowht{8pt}
        \textbf{Method}   & \textbf{TFLOPs ($\downarrow$)} & \textbf{AOKVQA} & \textbf{SQA}
 & \textbf{MME} & \textbf{POPE} & \textbf{MMB} & \textbf{MMB$\mathbf{^{CN}}$} & \textbf{LLaVA$\mathbf{^W}$} & \textbf{SEED$\mathbf{^I}$} \\
        \hline \xrowht{4pt}  
        LLaVA-1.5-7B~\cite{llava1-5}  & \multirow{2}{*}{11.05 (100\%)} & \multirow{2}{*}{77.8} & \multirow{2}{*}{70.8} & \multirow{2}{*}{1510} &  \multirow{2}{*}{86.1} & \multirow{2}{*}{65.3} & \multirow{2}{*}{59.4} & \multirow{2}{*}{65.5} & \multirow{2}{*}{66.7}  \\ 
        \, ~\textit{(T=576,L=32)} & & & & & & & & &   \\
        \hdashline \xrowht{4pt}
        + ShortGPT~\cite{shortgpt}$^*$   & 8.29 (75.0\%) & 74.2 & 64.6 & 964 &  69.7 & 50.4 & 37.5 & 52.9 & 54.3  \\ \xrowht{4pt}
        + RoE-LLaVA~\cite{RoE_LLaVA} & 8.24 (74.6\%) & - & 68.7 & - & - & 64.6 & - & - & 57.8 \\ \xrowht{4pt}
        + $\gamma$-MoD~\cite{luo2024gamma} & 
        7.98 (71.9\%) & - & 64.7 & 1342.1 & 86.0 & 59.3 & - & - & - \\
        \rowcolor[gray]{.95} \xrowht{4pt}
         + \modelname (\textit{T=256, L=28})  & 6.89 (62.4\%)  & 77.7 & \textbf{70.4} & \textbf{1351} & \textbf{80.6} & \textbf{62.6} & \textbf{54.5} & \textbf{64.8} & \textbf{65.2}    \\ 
         \rowcolor[gray]{.95} \xrowht{4pt} 
         + \modelname (\textit{T=176, L=28})$\dag$  & \textbf{6.19 (56.0\%)} &  \textbf{77.7} & 70.0 & 1300 & 76.1 & 62.2 & 53.8 & 63.8 & 62.7    \\ 
        \hdashline  \xrowht{4pt}
        + Random~\cite{llavolta}   & 5.93 (53.7\%) & 72.7 & 69.3 & 1142 &  55.8 & 39.7 & 33.3 & 47.6 & 52.2  \\  \xrowht{4pt}
  
        + LLaVolta~\cite{llavolta}    & 5.93 (53.7\%) & 74.9 & 69.4 & 1150 &  70.1 & 56.4 & 46.5 & 55.6 & 55.7   \\  \xrowht{4pt}
        + FastV~\cite{fastv}  & 5.93 (53.7\%) & 75.5 & 69.4 & \textbf{1298} & 65.6 &  60.1 & 53.0 & 54.8 & 56.3   \\
        \rowcolor[gray]{.95} \xrowht{4pt} 
        + \modelname (\textit{T=144, L=24})  & 5.18 (46.9\%) &  \textbf{77.6} & \textbf{69.8} & 1292 & \textbf{75.9} & \textbf{61.1} & \textbf{53.1} & \textbf{63.6} & \textbf{60.2}    \\ \rowcolor[gray]{.95} \xrowht{4pt} 
        + \modelname (\textit{T=128, L=24})  & \textbf{5.07 (45.9\%)} &  76.4 & 69.6 & 1286& 73.6 & 60.2 & 50.7 & 62.5 & 59.4    \\  
        \hline \xrowht{4pt} 
         Qwen-VL-Chat-9B~\cite{qwen-vl} & \multirow{2}{*}{10.01 (100\%)}  &\multirow{2}{*}{75.6} & \multirow{2}{*}{68.2} & \multirow{2}{*}{1487} & \multirow{2}{*}{86.5} & \multirow{2}{*}{60.6} & \multirow{2}{*}{56.7} & \multirow{2}{*}{73.5}  & \multirow{2}{*}{65.4} \\
         \qquad \textit{(T=256,L=32)} & & & & & & & & & \\
                 \hdashline \xrowht{4pt}
         + ShortGPT~\cite{shortgpt}$^*$   & 7.52 (75.1\%) & 63.5 & 52.6  & 1398 & 81.1 & 46.6 & 39.2 & 61.3 & 56.0  \\   \xrowht{4pt}
        + Random~\cite{llavolta}$^*$   & 7.21 (72.0\%) & 70.1 & 64.9 & 1138 &  80.2 & 44.3 & 37.6 & 62.5 & 57.7   \\  \xrowht{4pt}

        + LLaVolta~\cite{llavolta}$^*$  & 7.21 (72.0\%) &  71.6 & 65.3 & 1336 & 80.8 & 51.1 & 45.8 & 64.0 & 59.8  \\
         \xrowht{4pt}
        + FastV~\cite{fastv}$^*$  & 7.21 (72.0\%) &  72.2 & 65.9 & 1405 & 81.4 &  53.5 & 49.1 & 64.8 & 60.1  \\  \rowcolor[gray]{.95}        \xrowht{4pt}
        + \modelname (\textit{T=102, L=30})$\dag$  & 7.34 (73.3\%) & \textbf{74.7} & \textbf{66.8} & \textbf{1464} & \textbf{82.5} & \textbf{57.3} & \textbf{55.1} & \textbf{69.3} & \textbf{61.7}   \\ \rowcolor[gray]{.95} \xrowht{4pt}
        + \modelname (\textit{T=72, L=28})  & \textbf{6.61 (66.0\%)} &  73.5 & 66.3 & 1425 & 81.9 & 54.3 & 50.3 & 68.5 & 60.2    \\        
        \hline
    \end{tabular}

    \label{table:main_results}
\end{table*}

\begin{table}
    \centering
    \small
    \setlength{\tabcolsep}{6pt}
    \caption{Performance Comparison of different model settings.  $^*$ denotes reproduced results, and LLaVA-1.5-7B~\cite{llava1-5} is used as the backbone.}
    \begin{tabular}{l|lccc}
     \hline \xrowht{8pt}
        ~  &  \textbf{Method} &  \textbf{TFLOPs ($\downarrow$)} & \textbf{POPE} &  \textbf{SEED$\mathbf{^I}$}  \\
        \hline
        \rowcolor{mygray}
         & \textit{Token-level} & &  &  \\
        1  & Random Dropping & 5.93 (53.7\%) & 55.8 & 52.2 \\ 
        2  & LLaVolta~\cite{llavolta} & 5.93 (53.7\%) &  70.1 & 55.7 \\  
        3  & FastV~\cite{fastv} & 5.93 (53.7\%) & 65.6 & 56.3 \\  
         \hdashline
        4  & \textit{joint-then-token  } & 5.93 (53.7\%) & 71.3 & 57.9 \\  
        5  & \textit{token-router } & 5.93 (53.7\%) & 72.7 & 59.0 \\  
        \hline
        \rowcolor{mygray}
         & \textit{Layer-level} & &  &  \\
        6  & Random Dropping & 8.29 (75.0\%) &
        74.2 & 52.6 \\ 
        7  & ShortGPT~\cite{shortgpt}$^*$ & 8.29 (75.0\%) & 79.3 & 58.0 \\
         \hdashline
        8  & \textit{joint-then-layer } & 8.29 (75.0\%) & 82.1 & 60.5 \\  
        9  & \textit{layer-router } & 8.29 (75.0\%) & 82.9 & 61.1 \\  
        
        \hline
    \end{tabular}
    \label{table:single_level}
\end{table}

\begin{table}
    \centering
    \small
    \caption{\textbf{Ablation study.} We ablate key components to demonstrate the effectiveness of our method. LLaVA-1.5-7B~\cite{llava1-5} is used as the backbone, and ``Full'' denotes the original model. The ``Layer'', ``Token'' and ``Joint'' indicate layer skipping, token dropping and joint pruning, respectively. DPO denotes the DPO strategy, while ``Noise'' is the Gaussian Noise Perturbation.
        }
    \setlength{\tabcolsep}{1pt}
    \begin{tabular}{l|c|cccc|cc}
     \hline \xrowht{8pt}
        ~ & \textbf{TFLOPs}  & \textbf{Layer} & \textbf{Token} & \textbf{DPO} & \textbf{Noise}  & \textbf{POPE} &  \textbf{SEED$\mathbf{^I}$}  \\
        \hline
        \rowcolor{mygray}
        \xrowht{4pt} 
         ~& \textbf{\textit{Full (T=576,L=32)}} &  &  &  &   &  &  \\         \xrowht{4pt}  
        1 & 11.05 (100\%) &  &  &  &   & 86.1 & 66.7 \\
 
        \hdashline

        \rowcolor{mygray}
        \xrowht{4pt} 
         ~& \textbf{\textit{Layer (T=576,L=28)}} &  &  &  &   &  &  \\         \xrowht{4pt} 
        2 & 8.29 (75.0\%) & \checkmark &  &  &   & 81.4 & 61.6 \\
        \xrowht{4pt} 
        3 & 8.29 (75.0\%) & \checkmark &  & \checkmark &   & 85.8 & 65.1 \\
        \hdashline

        \rowcolor{mygray}
        \xrowht{4pt} 
         ~& \textbf{\textit{Token (T=176,L=32)}} &  &  &  &   &  &  \\         \xrowht{4pt} 
        4 & 6.96 (63.0\%) &  & \checkmark &  &   & 72.8 & 59.8 \\
        \xrowht{4pt} 
        5 & 6.96 (63.0\%) &  & \checkmark & \checkmark &  & 74.9 & 61.2 \\
        \xrowht{4pt} 
        6 & 6.96 (63.0\%)) &  & \checkmark &\checkmark & \checkmark  & 76.5 & 63.0 \\
        \hdashline
        \rowcolor{mygray}
        \xrowht{4pt}   
        ~& \textbf{\textit{Joint (T=176,L=28)}} &  &  &  &   &  &  \\         \xrowht{4pt} 
        7 & 6.19 (56.0\%) & \checkmark & \checkmark &  &   & 64.4 & 54.5 \\
        \xrowht{4pt}   
        8 & 6.19 (56.0\%) & \checkmark & \checkmark & \checkmark & \checkmark & 76.1 & 62.7 \\
        \hline

    \end{tabular}
    \label{tab:ablation}
\end{table}

\subsection{Experimental Settings}

\paragraph{Benchmarks and Backbones.}
We conduct extensive experiments on eight multimodal benchmarks, including AOKVQA~\cite{aokvqa}, Science-QA~\cite{scienceqa}, MME~\cite{mme}, POPE~\cite{pope}, MMBench~\cite{mmbench}, MMBench-CN\;~\cite{mmbench}, LLaVA-Bench-in-the-Wild~\cite{llava} and SEED-Bench-Image~\cite{seed_bench}. Moreover,
we apply \modelname into two prevalent LVLMs, including  LLaVA-v1.5-7B~\cite{llava1-5} and Qwen-VL-Chat-9B~\cite{qwen-vl}. During inference, we adopt the same settings as in FastV~\cite{fastv} (such as the systematic prompt) for fair comparisons.

\paragraph{Implementation Details.}
Following FastV~\cite{fastv} and LLaVolta~\cite{llavolta}, we report the theoretical TFLOPs and the corresponding reduction ratio.
The \modelname is trained on four NVIDIA 3090 GPUs and the batch size is 4.
For fair comparisons, we adopt the same settings as in FastV~\cite{fastv} such as systematic prompt and decoding strategy.
In our experiments, 5k data samples (less than 1\%) are randomly sampled from LLaVA-80k~\cite{llava} to construct preference data.
Note that we set the pruning actions $K=4$ and $M=400$. During testing, the $K$ and $M$ can also be assigned by users without retraining a new $\mathcal{O_\phi}$.

\subsection{Quantitative Evaluation}

\paragraph{Main Results.}

In Table \ref{table:main_results}, we compare the performance of our \modelname with previous acceleration methods on  two prevalent LVLMs (\textit{i.e.,} LLaVA-v1.5-7B~\cite{llava1-5} and Qwen-VL-Chat-9B~\cite{qwen-vl}). For LLaVA-v1.5~\cite{llava1-5}, we quote related results from~\cite{llavolta}.
Our meta-router is trained under the setting of removing 400 tokens and 4 layers, and this result is marked with $\dag$.
In Table \ref{table:main_results},  we use $T$ and $L$ to represent the number of retained tokens and layers for clarity. 
From the table, it can be observed that our \modelname obtains better performance across nine benchmarks. Specifically, the results show that our \modelname can outperform the state-of-the-art method FastV~\cite{fastv} by 3.3\% while achieving more compression ratio of approximately 6.8\%. In addition, when we further compress the model by only keeping 128 visual tokens and 24 layers, we observe that our method exceeds the SOTA~\cite{fastv} by 7.8\% lower TFLOPs. More importantly, compared to those training-based methods 
(\textit{e.g.,} RoE-LLaVA~\cite{RoE_LLaVA}, 
our \modelname still obtains the SOTA performance without updating the parameters of pruned LVLMs. Finally, we expand our approach to the Qwen-VL-Chat~\cite{qwen-vl}. The results show that the \modelname can be regarded as a plug-and-play module and conveniently transferred into other VL models. As a preliminary exploration, we also report different versions across various combinations of tokens and layers. Notably, \textbf{\textit{one-training is enough}}, these results can be directly obtained without retraining the meta-router. The flexibility allows the model to quickly adapt to different resource constraints and performance requirements. 

 \begin{figure*}[h]
    \centering
    \includegraphics[width=1\textwidth]{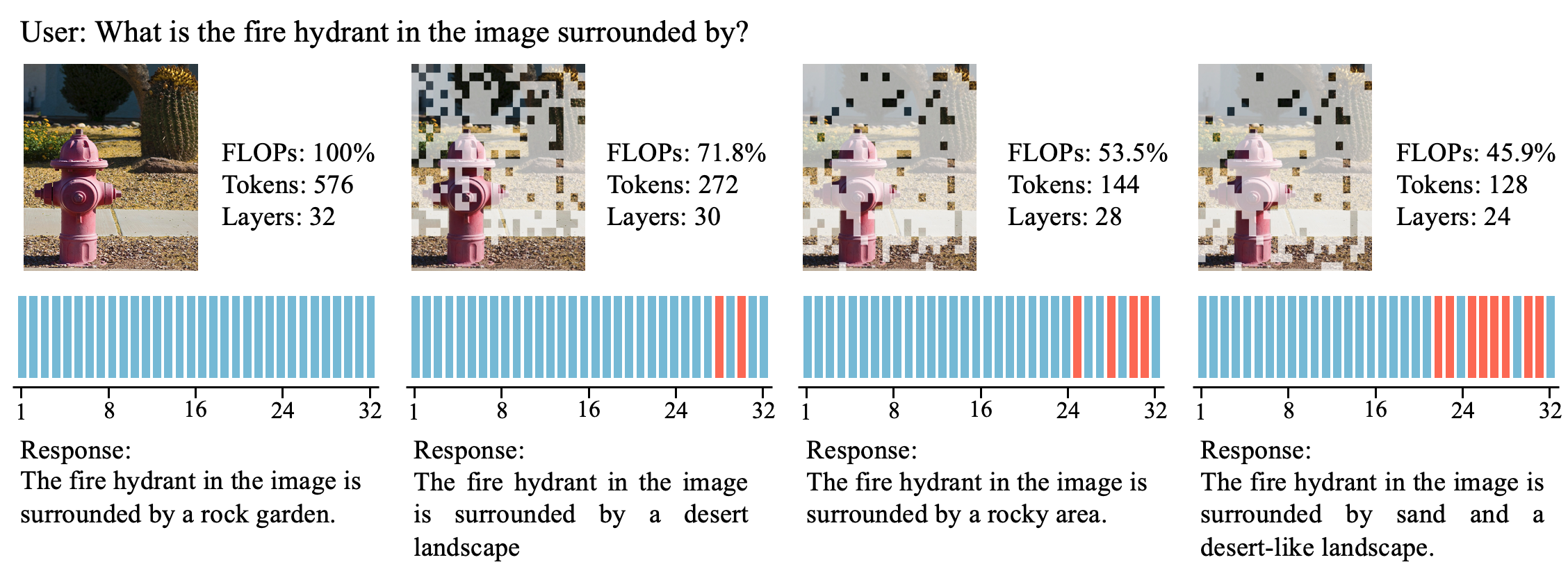}
    \caption{Qualitative results of our PAR method using LLaVA-1.5-7B~\cite{llava1-5}. The tokens and layers are gradually ablated to compress the model's FLOPs.}
\label{fig:qual_results}
\end{figure*}

\paragraph{Token or Layer Pruning.}
In Table~\ref{table:single_level}, we further conduct experiments on the LLaVA-v1.5-7B~\cite{llava1-5} to verify the effectiveness of our method with specific settings. 
Specifically, we first train a meta-router (\emph{i.e.,} \textit{T=176, L=28} in Table~\ref{table:main_results}). During inference, we only perform the token or layer compression branch, referred to as ``joint-then-token'' and ``joint-then-layer'', respectively. It can be observed that the \modelname still achieves outstanding performance when only performing single-branch compression.
Additionally, in the ``token-router'' and ``layer-router'', we retrain two specialized routers to execute layer pruning or token removal, respectively.
We observe that our \modelname exceeds these rule-based strategies~\cite{shortgpt,fastv} by a big margin. The above results show that the proposed \modelname has great flexibility to tackle different acceleration scenarios.

\subsection{Ablation Study}

As shown in Table~\ref{tab:ablation}, we conduct several ablation studies on the POPE~\cite{aokvqa} and SEED~\cite{seed_bench} to systematically discuss our method.
In the first row of Table \ref{tab:ablation}, we report the original results with LLaVA-v1.5-7B~\cite{llava1-5}. Moreover, we construct three groups of experiments to verify the effectiveness of different modules. Specifically, 
we first focus on the layer-level redundancy in 2-3 rows of this Table. 
In row 2, we skip the first four layers with the highest cosine similarity. Then, the DPO strategy is introduced to train a specific layer-router (\emph{i.e.,} the row 3). It can be observed that our method achieves better results compared to simply skipping specific layers.
In rows 4-6, we explore the token-level compression method. In row 4, we report corresponding results by dropping visual tokens using FastV~\cite{fastv}. Then the DPO and Gaussian Noise are incorporated to construct different token selection manners in rows 5-6. These results indicate that our method provides a more effective strategy for token pruning. 
Finally, we directly combine the rule-based layer skipping~\cite{shortgpt} and token pruning~\cite{fastv} in row 7. The result exhibits a significant decline due to the mutual influence between the token and layer. Conversely, our \modelname in row 8 can improve the simple combining method by 11.7 and 8.2 points on POPE and SEED.

\subsection{Qualitative Results}
\label{Qualitative Results}
As shown in Fig.\ref{fig:qual_results}, we visualize the model acceleration process on LLaVA-v1.5-7B~\cite{llava1-5} to further analyze and verify the proposed \modelname. 
In this figure, we progressively ablate visual tokens and layers to reduce the FLOPs ratio (from left to right).
From the figure, we find that our \modelname effectively mitigates computation costs while maintaining reasoning ability.
Notably, when \modelname removes $\sim$80\% (\emph{i.e.,} only remaining 128 tokens) of the visual tokens and skips 8 layers, resulting in only 45.9\% FLOPs ratio, the model is still able to follow the instruction and provide a reasonable response.  

%% file: sec/6_conclusion.tex
\section{Conclusion}
\label{sec:conclusion}

In this paper, we focus on reducing the computational cost of Large Vision-Language Models (LVLMs) across parameters and inputs perspectives. 
By building comprehensive comparisons, we empirically demonstrate that LVLMs suffer from serious redundancy across layers and tokens. 
Based on a series of meaningful insights, we propose a new Pruning All-Rounder (PAR) framework, which can jointly compress the model's parameters and inputs in a self-supervised learning manner. 
As a preliminary exploration, we expect our work to serve as a strong baseline and inspire our community to explore more effective acceleration strategies.

\clearpage

\section*{Acknowledgments}
This work is supported by the National Natural Science Foundation of China (No. U23B2013).